\documentclass[10pt,twocolumn,letterpaper]{article}

\usepackage{graphicx}
\usepackage{booktabs}
\usepackage{lscape}
 \usepackage{multirow}

\usepackage[pagenumbers]{main} 

\usepackage[dvipsnames]{xcolor}

\definecolor{cvprblue}{rgb}{0.21,0.49,0.74}
\usepackage[pagebackref,breaklinks,colorlinks,citecolor=cvprblue]{hyperref}

\def\paperID{*****} 
\def\confName{CVPR}
\def\confYear{2024}

\title{T2I-FineEval: Fine-Grained Compositional Metric for Text-to-Image Evaluation}

\author{
    \textit{Seyed Mohammad Hadi Hosseini}\thanks{Equal contribution} \quad
    \textit{Amir Mohammad Izadi}\footnotemark[1] \quad
    \textit{Ali Abdollahi} \\
    \textit{Armin Saghafian} \quad
    \textit{Mahdieh Soleymani Baghshah} \\
    Sharif University of Technology \\
    {\tt\small \{hadi.hosseini17, amirmohammad.izadi01, soleymani\}@sharif.edu}\\
        {\tt\small \{armin.saghafian, aliabdollahi024a\}@gmail.com} 
}

\begin{document}
\maketitle
\begin{abstract}
Although recent text-to-image generative models have achieved impressive performance, they still often struggle with capturing the compositional complexities of prompts including attribute binding, and spatial relationships between different entities. This misalignment is not revealed by common evaluation metrics such as CLIPScore. Recent works have proposed evaluation metrics that utilize Visual Question Answering (VQA) by decomposing prompts into questions about the generated image for more robust compositional evaluation. 
Although these methods align better with human evaluations, they still fail to fully cover the compositionality within the image. To address this, we propose a novel metric that breaks down images into components, and texts into fine-grained questions about the generated image for evaluation. Our method outperforms previous state-of-the-art metrics, demonstrating its effectiveness in evaluating text-to-image generative models. Code is available at \url{https://github.com/hadi-hosseini/T2I-FineEval}.

\end{abstract}    
\section{Introduction}
\label{sec:intro}

\begin{figure*}[ht!]
  \centering
  \label{Matching Method}
  \includegraphics[width=\textwidth]{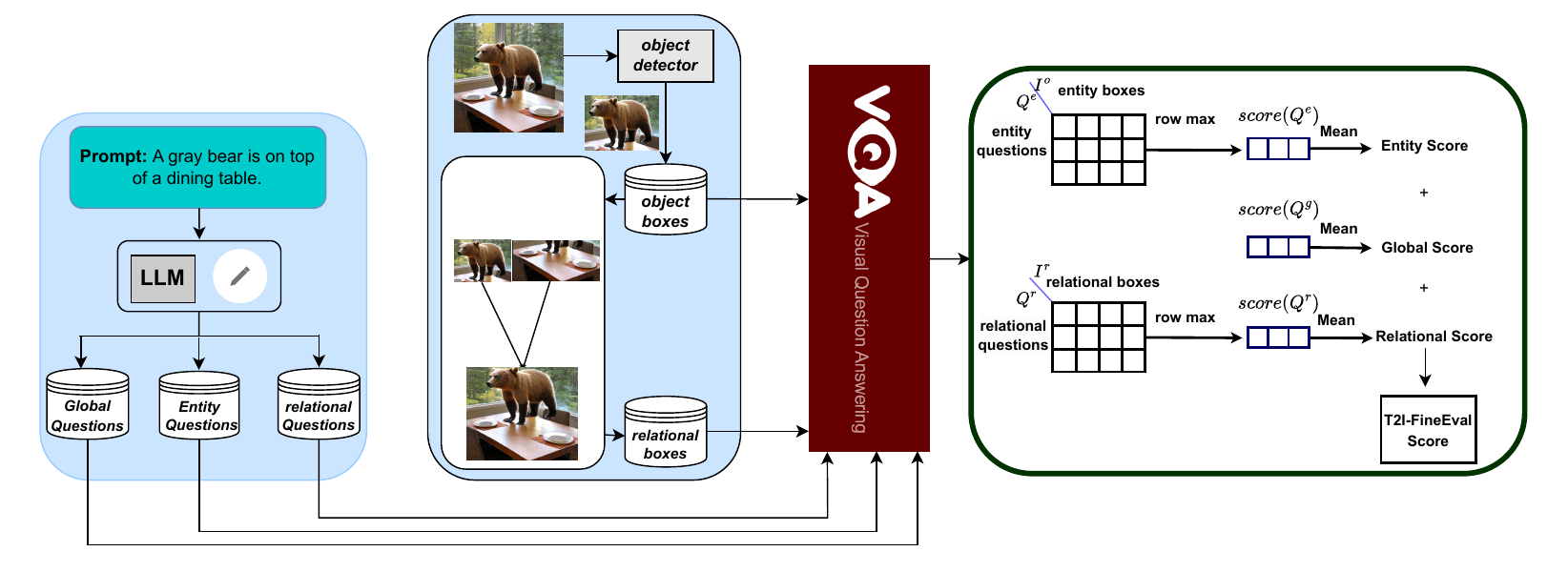}
  \caption{
  \textbf{Illustration of T2I-FineEval:}
  A Large Language Model generates coarse-grained, entity-specific, and relational questions from a text prompt. An object detector identifies entities in an image created from the same prompt, forming boxes for each entity and pairing them to capture candidate relationships (relational boxes).
  A Visual Question-Answering (VQA) model is used to ask coarse-grained questions about the entire image, resulting in a coarse-grained score. Entity-specific questions are matched with their corresponding entity boxes, while relational questions are paired with relational boxes, forming a question-box matrix. The highest scoring box for each question is selected, and these scores contribute to an image's fine-grained score. The overall score is then the average of the fine-grained and coarse-grained scores.
  }
\end{figure*}

Recent text-to-image generative models have shown a remarkable ability to create desired scenes based on prompts. However, despite the recent advancements, these models often struggle to accurately generate images from complex prompts, particularly when the prompts involve intricate relationships between components \cite{liu2022compositional}, multiple components, and attribute binding \cite{feng2022training,liu2022compositional,petsiuk2022human}. These issues frequently manifest as attributes being incorrectly assigned to objects, or as an overall incoherence in the resulting scene. Due to these weaknesses, developing metrics that align well with human ratings has become a crucial issue. 
CLIPScore \cite{hessel2021clipscore} measures the alignment between text and generated images via the similarity between text and image embeddings. However, it does not perform well in assessing generated images when the prompt is complex. An alternative approach used in some recent works involves using captioning models to generate a caption for the generated image and then evaluating the similarity between the initial prompt and the generated caption, as seen with CIDEr \cite{vedantam2015cider} and SPICE \cite{anderson2016spice}. 
Unfortunately, these evaluation methods are not ideal because the generated caption is based on the image as a whole, is prone to overlooking noticeable regions of the image, or focusing on non-essential regions \cite{kasai2021transparent}. 

Recently, several metrics have been introduced to evaluate the alignment between text and generated images using pretrained VQA models such as mPLUGlarge \cite{li2205mplug} and BLIP2 \cite{li2023blip}. Among these metrics, some, like TIFA \cite{hu2023tifa}, CompBench  metric \cite{huang2023t2i}, and DA-Score \cite{singh2023divide}, aim to capture compositional complexities. These metrics break the text into fine-grained questions to evaluate the generated image, thereby achieving higher alignment with human evaluation. The primary drawback of these metrics is that, by focusing solely on the main image and asking fine-grained questions, they risk neglecting vital aspects of evaluation—such as attribute binding and spatial relationships between entities—due to their inherent limitations in compositional understanding.
We further improve the existing approaches by decomposing the image into their components too. 

We propose an evaluation method called T2I-FineEval (Text-to-Image Fine-grained Evaluation) to address the limitations of previous compositional metrics. T2I-FineEval breaks down images into entity and relational components, using object detector to extract entity components, and then merge them into relational components. Furthermore, we generate questions based on the given text, and these questions along with image components are fed into a VQA model to assign a score to each of the question-component pairs. Ultimately, we incorporate a matching method similar to FILIP matching \cite{yao2021filip} to determine the final score, effectively aligning questions with image components for a more robust evaluation.
Our contribution consists of two main parts: (1) We proposed a fine-grained evaluation of text and image, breaking down images into entity and relational components, and extracting entity and relational questions from the text. (2) We utilize fine-grained text-to-image matching between questions and image components along with the coarse-grained score to compute a final score that closely aligns with human evaluation.
\section{Method}

We introduce a method to assess how well an image aligns with its corresponding text. Our approach evaluates compositional similarity by focusing on attribute binding and interactions among multiple objects. To achieve this, we break the text into separate assertions and make a set of questions based on the single components (entity questions), as well as questions that address the relationships among multiple components (relational questions), capturing the interactions of these components. Moreover, we generate general questions from the original text and assess its alignment with the whole image. As for the image, we use an object detector to extract object boxes from the image, which we then merge each pair of object boxes to create candidate relational boxes. Finally, we match the entity questions with object boxes, the relational questions with relational boxes, and the coarse-grained questions with the original image to compute the final similarity score.
\\
\textbf{Computing Compositional Score.}
Given an image $\mathcal{I}$ and a prompt $\mathcal{P}$, we first use GPT-4 \cite{unknown} to decompose $\mathcal{P}$ into separate assertions. We then ask GPT-4 to generate a series of entity questions $\{Q_i^e\}_{i=1}^{n_e}$ from each assertion and a set of relational questions $\{Q_i^r\}_{i=1}^{n_r}$ regarding the interactions between entities. Additionally, we make a series of general questions $\{Q_i^g\}_{i=1}^{n_g}$ based on the original prompt.
Concerning the image, we use YOLOv9 \cite{wang2024yolov9} to extract object boxes, denoted as $\{I_i^o\}$. We then merge these boxes in pairs to create candidate relational boxes that capture interactions between objects, denoted as $\{I_i^r\}$.
To compute the final score, we assign a score to each question based on the components we extracted with the object detector, demonstrating how well each image component answers each question. We use the pretrained BLIP-VQA \cite{li2022blip} model (a BLIP model fine-tuned for visual question answering) to evaluate. Each question-box pair is scored by considering the probability of "YES" that BLIP-VQA provides when the question is asked about that image component.
We assess how well a question has been answered based on its alignment with the box that yields the highest score. 
To maintain consistency, we only pair questions and image components within their respective groups, i.e. entity questions with object boxes, relational questions with relational boxes, and coarse-grained questions with the entire image.
\begin{table*}[ht]
    \centering
    \resizebox{\textwidth}{!}{
    \begin{tabular}{lcccccccccccccccc}
        \toprule
        & \multicolumn{8}{c}{\textbf{Shape}} & \multicolumn{8}{c}{\textbf{Color}} \\
        \cmidrule(lr){2-9} \cmidrule(lr){10-17}
        \textbf{Model} & \multicolumn{2}{c}{\textbf{B-VQA}} & \multicolumn{2}{c}{\textbf{TIFA}} & \multicolumn{2}{c}{\textbf{DA-Score}} & \multicolumn{2}{c}{\textbf{T2I-FineEval}} & \multicolumn{2}{c}{\textbf{B-VQA}} & \multicolumn{2}{c}{\textbf{TIFA}} & \multicolumn{2}{c}{\textbf{DA-Score}} & \multicolumn{2}{c}{\textbf{T2I-FineEval}} \\
        \cmidrule(lr){2-3} \cmidrule(lr){4-5} \cmidrule(lr){6-7} \cmidrule(lr){8-9} \cmidrule(lr){10-11} \cmidrule(lr){12-13} \cmidrule(lr){14-15} \cmidrule(lr){16-17}
        & \textbf{$\tau$} & \textbf{$\rho$} & \textbf{$\tau$} & \textbf{$\rho$} & \textbf{$\tau$} & \textbf{$\rho$} & \textbf{$\tau$} & \textbf{$\rho$} & \textbf{$\tau$} & \textbf{$\rho$} & \textbf{$\tau$} & \textbf{$\rho$} & \textbf{$\tau$} & \textbf{$\rho$} & \textbf{$\tau$} & \textbf{$\rho$} \\
        \midrule
        Stable v1.4 \cite{rombach2022high} & 0.24 & 0.32 & 0.20 & 0.27 & 0.36 & 0.48 & 0.41 & 0.54 & 0.75 & 0.88 & 0.22 & 0.28 & 0.69 & 0.83 & 0.71 & 0.84 \\
        Stable v2 \cite{rombach2022high} & 0.27 & 0.38 & 0.27 & 0.35 & 0.26 & 0.36 & 0.32 & 0.46 & 0.63 & 0.80 & 0.44 & 0.57 & 0.64 & 0.79 & 0.68 & 0.83 \\
        Composable v2 \cite{liu2022compositional} & 0.16 & 0.23 & 0.37 & 0.48 & 0.17 & 0.27 & 0.36 & 0.47 & 0.60 & 0.79 & 0.35 & 0.44 & 0.59 & 0.77 & 0.66 & 0.83 \\
        Structure v2 \cite{feng2022training} & 0.29 & 0.41 & 0.12 & 0.19 & 0.35 & 0.49 & 0.48 & 0.65 & 0.62 & 0.78 & 0.32 & 0.41 & 0.55 & 0.72 & 0.62 & 0.78 \\
        Attn-Exct v2 \cite{chefer2023attend} & 0.38 & 0.52 & 0.22 & 0.28 & 0.41 & 0.55 & 0.47 & 0.63 & 0.59 & 0.73 & 0.28 & 0.37 & 0.46 & 0.59 & 0.62 & 0.77 \\
        GORS \cite{huang2023t2i} & 0.27 & 0.36 & 0.21 & 0.28 & 0.29 & 0.39 & 0.37 & 0.51 & 0.50 & 0.66 & 0.49 & 0.60 & 0.39 & 0.51 & 0.52 & 0.68 \\
        \midrule
        \textbf{Mean} & 0.27 & 0.37 & 0.23 & 0.31 & 0.31 & 0.42 & \textbf{0.40} & \textbf{0.54} & 0.62 & 0.77 & 0.35 & 0.45 & 0.55 & 0.70 & \textbf{0.64} & \textbf{0.79} \\
        \bottomrule
    \end{tabular}
    }
    \caption{The performance of various automated evaluation metrics on images generated by different text-to-image (T2I) generative models using T2I-CompBench benchmark in the color and shape categories. The last row represents the average performance across all models.}
    \label{tab:performance_comparison1}
\end{table*}
\begin{table*}[t]
    \centering
    \resizebox{\textwidth}{!}{
    \begin{tabular}{lcccccccccccccccc}
        \toprule
        & \multicolumn{8}{c}{\textbf{Texture}} & \multicolumn{8}{c}{\textbf{Spatial Relation}} \\
        \cmidrule(lr){2-9} \cmidrule(lr){10-17}
        \textbf{Model} & \multicolumn{2}{c}{\textbf{B-VQA}} & \multicolumn{2}{c}{\textbf{TIFA}} & \multicolumn{2}{c}{\textbf{DA-Score}} & \multicolumn{2}{c}{\textbf{T2I-FineEval}} & \multicolumn{2}{c}{\textbf{UniDet}} & \multicolumn{2}{c}{\textbf{TIFA}} & \multicolumn{2}{c}{\textbf{DA-Score}} & \multicolumn{2}{c}{\textbf{T2I-FineEval}} \\
        \cmidrule(lr){2-3} \cmidrule(lr){4-5} \cmidrule(lr){6-7} \cmidrule(lr){8-9} \cmidrule(lr){10-11} \cmidrule(lr){12-13} \cmidrule(lr){14-15} \cmidrule(lr){16-17}
        & \textbf{$\tau$} & \textbf{$\rho$} & \textbf{$\tau$} & \textbf{$\rho$} & \textbf{$\tau$} & \textbf{$\rho$} & \textbf{$\tau$} & \textbf{$\rho$} & \textbf{$\tau$} & \textbf{$\rho$} & \textbf{$\tau$} & \textbf{$\rho$} & \textbf{$\tau$} & \textbf{$\rho$} & \textbf{$\tau$} & \textbf{$\rho$} \\
        \midrule
        Stable v1.4 & 0.54 & 0.72 & 0.35 & 0.46 & 0.57 & 0.74 & 0.56 & 0.74 & 0.54 & 0.59 & 0.13 & 0.16 & 0.21 & 0.27 & 0.33 & 0.42 \\
        Stable v2 & 0.44 & 0.61 & 0.37 & 0.48 & 0.44 & 0.61 & 0.45 & 0.62 & 0.4 & 0.41 & 0.17 & 0.2 & 0.14 & 0.19 & 0.31 & 0.39 \\
        Composable v2 & 0.27 & 0.36 & 0.29 & 0.38 & 0.25 & 0.34 & 0.22 & 0.29 & 0.52 & 0.55 & 0.34 & 0.42 & 0.35 & 0.44 & 0.38 & 0.47 \\
        Structure v2 & 0.5 & 0.67 & 0.35 & 0.46 & 0.49 & 0.64 & 0.47 & 0.62 & 0.43 & 0.45 & 0.14 & 0.17 & 0.18 & 0.23 & 0.39 & 0.5 \\
        Attn-Exct v2 & 0.53 & 0.7 & 0.46 & 0.62 & 0.54 & 0.69 & 0.55 & 0.71 & 0.41 & 0.45 & 0.2 & 0.23 & 0.35 & 0.45 & 0.32 & 0.41 \\
        GORS & 0.44 & 0.6 & 0.38 & 0.52 & 0.44 & 0.58 & 0.47 & 0.61 & 0.52 & 0.58 & 0.31 & 0.37 & 0.31 & 0.42 & 0.4 & 0.52 \\
        \midrule
        \textbf{Mean} & 0.45 & 0.61 & 0.37 & 0.49 & \textbf{0.46} & \textbf{0.6} & 0.45 & \textbf{0.6} & \textbf{0.47} & \textbf{0.5} & 0.21 & 0.26 & 0.26 & 0.33 & 0.36 & 0.45 \\
        \bottomrule
    \end{tabular}
    }
    \caption{The performance of various automated evaluation metrics on images generated by different text-to-image (T2I) generative models using T2I-CompBench benchmark in the texture and spatial relation categories. The last row represents the average performance across all models.}
    \label{tab:performance_comparison2}
\end{table*}
\begin{equation}
\footnotesize
\label{eq:eq1}
    \text{score}(Q_i^{e}) =
    \max_j P (\text{"YES"} \mid Q_i^{e}, I_j^{o})
\end{equation}

\begin{equation}
\label{eq:eq2}
\footnotesize
    \text{score}(Q_i^{r}) =
    \max_j P (\text{"YES"} \mid Q_i^{r}, I_j^{r})
\end{equation}

\begin{equation}
\footnotesize
    \text{score}(Q_i^{g}) = P (\text{"YES"} \mid Q_i^{g}, \mathcal{I})
\end{equation}
The final score of the original image and prompt $(\mathcal{I}, \mathcal{P})$ is then determined as the mean score of all the questions.

\begin{equation}
\footnotesize
    \text{fine-grained score}(\mathcal{I}, \mathcal{P}) =
    \frac{\sum_{i=1}^{n_e} \text{score}(Q_i^{e})}{2n_e} +
    \frac{\sum_{i=1}^{n_r} \text{score}(Q_i^{r})}{2n_r}
\end{equation}

\begin{equation}
\footnotesize
    \text{coarse-grained score} (\mathcal{I}, \mathcal{P}) = \frac{\sum_{i=1}^{n_g} \text{score}(Q_i^{g})}{n_g} 
\end{equation}


\begin{equation}
\footnotesize
    \text{score}(\mathcal{I},  \mathcal{P}) =
    \frac{\text{fine-grained score}(\mathcal{I}, \mathcal{P}) + \text{coarse-grained score} (\mathcal{I}, \mathcal{P})}{2}
\end{equation}

\begin{table*}[t]
    \centering
    \resizebox{\textwidth}{!}{
    \begin{tabular}{lcccccccccccccccc}
        \toprule
        & \multicolumn{8}{c}{\textbf{Non-spatial Relation}} & \multicolumn{8}{c}{\textbf{Complex}} \\
        \cmidrule(lr){2-9} \cmidrule(lr){10-17}
        \textbf{Model} & \multicolumn{2}{c}{\textbf{CLIPScore}} & \multicolumn{2}{c}{\textbf{TIFA}} & \multicolumn{2}{c}{\textbf{DA-Score}} & \multicolumn{2}{c}{\textbf{T2I-FineEval}} & \multicolumn{2}{c}{\textbf{3-in-1}} & \multicolumn{2}{c}{\textbf{TIFA}} & \multicolumn{2}{c}{\textbf{DA-Score}} & \multicolumn{2}{c}{\textbf{T2I-FineEval}} \\
        \cmidrule(lr){2-3} \cmidrule(lr){4-5} \cmidrule(lr){6-7} \cmidrule(lr){8-9} \cmidrule(lr){10-11} \cmidrule(lr){12-13} \cmidrule(lr){14-15} \cmidrule(lr){16-17}
        & \textbf{$\tau$} & \textbf{$\rho$} & \textbf{$\tau$} & \textbf{$\rho$} & \textbf{$\tau$} & \textbf{$\rho$} & \textbf{$\tau$} & \textbf{$\rho$} & \textbf{$\tau$} & \textbf{$\rho$} & \textbf{$\tau$} & \textbf{$\rho$} & \textbf{$\tau$} & \textbf{$\rho$} & \textbf{$\tau$} & \textbf{$\rho$} \\
        \midrule
        Stable v1.4 & 0.35 & 0.44 & 0.21 & 0.24 & 0.48 & 0.59 & 0.46 & 0.56 & 0.31 & 0.44 & 0.45 & 0.6 & 0.46 & 0.6 & 0.49 & 0.64 \\
        Stable v2 & 0.03 & 0.03 & -0.05 & -0.05 & 0.15 & 0.19 & 0.13 & 0.16 & 0.25 & 0.32 & 0.36 & 0.47 & 0.32 & 0.42 & 0.31 & 0.42 \\
        Composable v2 & 0.44 & 0.58 & 0.42 & 0.53 & 0.51 & 0.68 & 0.46 & 0.62 & 0.3 & 0.42 & 0.35 & 0.47 & 0.34 & 0.48 & 0.35 & 0.47 \\
        Structure v2 & -0.1 & -0.12 & -0.07 & -0.09 & 0.06 & 0.07 & 0.03 & 0.04 & 0.26 & 0.34 & 0.42 & 0.52 & 0.34 & 0.45 & 0.26 & 0.36 \\
        Attn-Exct v2 & 0.23 & 0.3 & 0.13 & 0.17 & 0.34 & 0.44 & 0.43 & 0.54 & 0.18 & 0.24 & 0.16 & 0.21 & 0.34 & 0.48 & 0.38 & 0.5 \\
        GORS & 0.11 & 0.13 & 0.21 & 0.22 & 0.25 & 0.31 & 0.28 & 0.34 & 0.25 & 0.34 & 0.34 & 0.43 & 0.32 & 0.42 & 0.39 & 0.5 \\
        \midrule
        \textbf{Mean} & 0.18 & 0.23 & 0.14 & 0.17 & \textbf{0.3} & \textbf{0.38} & \textbf{0.3} & \textbf{0.38} & 0.26 & 0.35 & 0.35 & 0.45 & 0.35 & \textbf{0.48} & \textbf{0.36} & \textbf{0.48} \\
        \bottomrule
    \end{tabular}
    }
    \caption{The performance of various automated evaluation metrics on images generated by different text-to-image (T2I) generative models using T2I-CompBench benchmark in the complex and Non-spatial rel categories. The last row represents the average performance across all models.}
    \label{tab:performance_comparison3}
\end{table*}

\section{Experiments}
\label{sec:exp}

In this section, we evaluate our proposed metric, T2I-FineEval, in comparison with several existing metrics to assess their effectiveness in measuring text-to-image generation quality. We conduct this evaluation across multiple text-to-image generative models using the T2I-CompBench benchmark \cite{huang2023t2i}. Our results demonstrate that the compositional approach of T2I-FineEval achieves superior alignment with human evaluation scores, highlighting its potential as a more reliable assessment metric.

\subsection{Evaluated models and Metrics}
\label{sec:evalMet}
We evaluated the performance of our metric, T2I-FineEval, DA-Score \cite{singh2023divide}, TIFA \cite{hu2023tifa}, and CompBench metric \cite{huang2023t2i} (B-VQA, UniDet \cite{zhou2022simple}, CLIPScore, 3-in-1) on six T2I generative models (Stable-Diffusion v1.4, Stable-Diffusion v2 \cite{rombach2022high}, Composable-Diffusion \cite{liu2022compositional}, Structure-Diffusion \cite{feng2022training}, and Attend-and-Excite \cite{chefer2023attend}) using the T2I-CompBench benchmark \cite{huang2023t2i}. 

Here, we explain the details of the metrics we used in our experiments.
The CompBench metric \cite{huang2023t2i} employs UniDet for assessing spatial relationships, B-VQA for attribute binding, and CLIPScore for evaluating non-spatial relationships. For complex prompts, a combination of these three metrics is used to cover, called 3-in-1.
In the TIFA score \cite{hu2023tifa}, $N$ multiple-choice question-answer pairs are generated from the text. Each question has one correct answer and if the most probable output among these choices is the correct one, a positive point is rewarded. The DA-Score \cite{singh2023divide} uses BLIP-VQA for score calculation. DA-Score uses GPT-4 and two human-generated examples to decompose each text into $N$ assertions and subsequently convert each assertion into a question for VQA.

\subsection{Benchmarks}
We evaluated the aforementioned metrics on T2I-CompBench which has three different categories, split into six sub-categories of compositionality: attribute binding (including three sub-categories: color, shape, and texture), object relationships (including two sub-categories: spatial relationships and non-spatial relationships), and complex compositions. Each sub-category in the benchmark contains 300 test prompts for the evaluation of compositional image generation.

\subsection{Implementation Details}

We evaluated our proposed metric, T2I-FineEval, on BLIP-VQA, and 
calculated the DA-Score using the same model.
To reproduce the TIFA score's results, we employed BLIP2, which is more powerful than BLIP-VQA. To calculate the T2I-CompBench score, we computed each sub-category separately as mentioned in section \ref{sec:evalMet}. To decompose the text into entity, relational, and global questions, we used GPT-4, taking advantage of In-Context Learning (ICL) with a two-shot sample in the prompt. To decompose each generated image into its components, we utilized the YOLOv9 \cite{wang2024yolov9} object detector. All experiments were carried out using a single RTX 3090 graphics card.

\subsection{Human Correlation of the Evaluation Metrics}

In our analysis, we calculate Kendall’s Tau ($\tau$) and Spearman’s Rho ($\rho$) for each evaluation metric to measure the correlation between the evaluated scores and human evaluation. Higher values of ($\tau$) and ($\rho$) indicate a better alignment with human evaluation. Both human evaluation and the other metrics are normalized between 0 and 1. Our study demonstrates that our proposed method has a stronger correlation with human evaluation compared to other metrics.

\subsection{Result Analysis}

We report our results on the T2I-CompBench benchmark, which include attribute binding (shape, color) (Table \ref{tab:performance_comparison1}), attribute binding (texture), spatial relationship (Table \ref{tab:performance_comparison2}), non-spatial relationship, and complex compositions (Table \ref{tab:performance_comparison3}). We report Kendall's Tau and Spearman's Rho correlation scores across various models, along with the average score. The proposed method outperforms the existing text-to-image alignment metrics in three categories (shape, color, and complex), achieving competitive results on the other two (texture and non-spatial relation). Table \ref{tab:performance_comparison4} displays the average score over all tasks,
where we demonstrate clear superiority over all models.

\begin{table}[t]
    \centering
    \resizebox{\columnwidth}{!}{
    \begin{tabular}{lcccccccc}
        \toprule
        \textbf{Model} & \multicolumn{2}{c}{\textbf{CompBench}} & \multicolumn{2}{c}{\textbf{TIFA}} & \multicolumn{2}{c}{\textbf{DA-Score}} & \multicolumn{2}{c}{\textbf{T2I-FineEval}} \\
        \cmidrule(lr){2-3} \cmidrule(lr){4-5} \cmidrule(lr){6-7} \cmidrule(lr){8-9}
        & \textbf{$\tau$} & \textbf{$\rho$} & \textbf{$\tau$} & \textbf{$\rho$} & \textbf{$\tau$} & \textbf{$\rho$} & \textbf{$\tau$} & \textbf{$\rho$} \\
        \midrule
        Stable v1.4 & 0.46 & 0.57 & 0.26 & 0.34 & 0.46 & 0.59 & \textbf{0.49} & \textbf{0.62} \\
        Stable v2 & 0.34 & 0.43 & 0.26 & 0.34 & 0.33 & 0.43 & \textbf{0.37} & \textbf{0.48} \\
        Composable v2 & 0.38 & 0.49 & 0.35 & 0.45 & 0.37 & 0.50 & \textbf{0.41} & \textbf{0.53} \\
        Structure v2 & 0.33 & 0.42 & 0.21 & 0.28 & 0.33 & 0.43 & \textbf{0.38} & \textbf{0.49} \\
        Attn-Exct v2 & 0.39 & 0.49 & 0.24 & 0.31 & 0.41 & 0.53 & \textbf{0.46} & \textbf{0.59} \\
        GORS & 0.35 & 0.45 & 0.32 & 0.40 & 0.33 & 0.44 & \textbf{0.41} & \textbf{0.53} \\
        \bottomrule
    \end{tabular}
    }
    \caption{The average performance across various automated evaluation metrics for all categories using the T2I-CompBench benchmark.}
    \label{tab:performance_comparison4}
\end{table}

\section{Conclusion}
We present T2I-FineEval, a fine-grained metric for alignment evaluation in generative text-to-image models, which takes advantage of decomposing both text and image. Unlike prior metrics that only compose the prompt and do not decompose images, ComTie can break both the image and the prompt into compositional components and evaluate them on a more fine-grained level. Applying our method can result in a better alignment with human evaluation than other state-of-the-art metrics.
To support our claims, we conducted a quantitative comparison with CompBench (B-VQA, UniDet, CLIPScore), TIFA, and DA-Score metrics on various text-to-image models and get better average results over all metrics.
{
    \small
    \bibliographystyle{main}
    \bibliography{main}
}


\end{document}